\documentclass[lettersize,journal]{IEEEtran}
\usepackage{amsmath,amsfonts}
\usepackage{algorithmic}
\usepackage{algorithm}
\usepackage{array}
\usepackage[caption=false,font=normalsize,labelfont=sf,textfont=sf]{subfig}
\usepackage{textcomp}
\usepackage{stfloats}
\usepackage{url}
\usepackage{verbatim}
\usepackage{graphicx}
\usepackage{cite}
\usepackage{threeparttable}
\usepackage{booktabs} 
\usepackage{multirow}
\usepackage{makecell}
\usepackage{booktabs}
\usepackage{threeparttable}
\begin{document}

\title{DBINDS - Can Initial Noise from Diffusion Model Inversion Help Reveal AI-Generated Videos? }

\author{Yanlin~Wu,
	Xiaogang~Yuan,
	Dezhi~An
\thanks{Yanlin Wu, Xiaogang Yuan, and Dezhi An are with the School of Cyber Security, 
	Gansu University of Political Science and Law, Lanzhou, Gansu, China.}
\thanks{Corresponding author: Xiaogang Yuan (email: yxg7349@gsupl.edu.cn).}}



\maketitle

\begin{center}
	{\small
		\textbf{Preprint Notice:} This work has been submitted to \textit{IEEE Transactions on Dependable and Secure Computing (TDSC)} for possible publication.\\
		This version is made available for research and academic discussion purposes only.}
\end{center}
\vspace{1em}

\begin{abstract}
AI-generated video has advanced rapidly, posing serious challenges to content security and forensic analysis. Existing detection methods primarily rely on pixel-level visual features and exhibit limited generalization to unseen generators. We propose DBINDS, a diffusion-model-inversion–based detection framework that, for the first time, extends analysis from the pixel domain to the latent space. We observe that the initial noise sequences obtained via diffusion model inversion from real and AI‑generated videos display markedly different variation patterns. Building on this observation, DBINDS constructs the Initial Noise Difference Sequence (INDS) as the core feature representation and introduces a multidimensional, multiscale analytical framework to comprehensively examine INDS for AI‑generated-video detection. Through systematic, experiment-driven exploration with feature-optimization strategies, we thoroughly analyze the latent feature space of INDS and find that a composite of spatiotemporal correlation and spatiotemporal texture features achieves strong zero-shot generalization and high detection accuracy. Importantly, the method retains notable advantages under limited training data: training on a single generator suffices to build a detector with strong cross-generator generalization, which is of practical value when labeled data are scarce or costly. Using Bayesian hyperparameter optimization and a LightGBM classifier, we validate DBINDS on the GenVid dataset. Trained on data from a single generator, DBINDS attains 78.08\% overall accuracy across seven unseen generators and one unseen real-video set, confirming its effectiveness under constrained training conditions and demonstrating robustness and cross-model transferability. These results establish INDS as an effective latent cue and provide a feasible path—with optimized feature combinations and best practices—for deployment in resource-constrained settings.
\end{abstract}


\section{Introduction}
\IEEEPARstart{A}{I}-driven video generation—epitomized by the recent advent of Sora \cite{res1,res2}—has advanced rapidly, achieving high visual fidelity, temporal coherence, and perceptual realism. While these capabilities broaden application prospects, they also lower the barriers to synthetic‑media manipulation, heightening risks of disinformation, evidentiary fabrication, and identity fraud\cite{kaur2024deepfake,10543910}. During the 2025 India–Pakistan conflict, for instance, AI‑generated images and videos circulated widely on social platforms, misleading the general public as well as mainstream news outlets and public officials, thereby distorting public perception and amplifying anxiety. These security concerns have encouraged more cautious release practices for state‑of‑the‑art generative models and slowed progress in related open‑source efforts. Accordingly, developing robust and reliable detectors for AI‑generated videos has become an urgent research priority\cite{cooke2025goodcointosshuman}.

Contemporary AI‑generated‑video detectors face three interrelated limitations. First, cross‑model generalization remains a principal challenge. Wang et al.\cite{res3} report that spatial artifacts vary widely across generative architectures and lack a stable signature; controlled experiments in the same study show that detectors readily overfit to generator‑specific cues, with sharp performance drops on previously unseen models. Given these pronounced discrepancies, building spatial‑feature–based detectors that remain valid across diverse architectures is difficult, and zero‑shot transferability to emergent generators such as Sora remains limited.

Second, feature‑learning–based approaches exhibit inherent constraints. Modern generators support diverse conditioning paradigms (e.g., text‑to‑video, image‑to‑video), affording broad creative latitude and yielding content of high structural and stylistic complexity. Much of the deepfake (facial‑manipulation) detection literature\cite{NEURIPS2024_b7d9b1d4,chen2022self,yu2024diff,Stamnas_2025_WACV} is tailored to faces and therefore fails on non‑facial or non‑human content\cite{res4,Kundu_2025_CVPR}. Moreover, most detectors rely primarily on visual cues to identify quality imperfections in generated content while overlooking mechanistic differences among generative pipelines. Such imperfections often manifest as spatiotemporal inconsistencies, including geometric and pose/motion mismatches\cite{chang2024mattersdetectingaigeneratedvideos,AIGVDet} and reduced temporal coherence\cite{he2024exposingaigeneratedvideosbenchmark}.

Finally, computational cost and deployability pose practical impediments. State‑of‑the‑art systems—such as the large‑parameter DeMamba architecture\cite{res5} and BusterX\cite{wen}, which builds on multimodal large language models with reinforcement learning—achieve strong accuracy but require extensive training corpora and substantial computation, constraining deployment in resource‑limited settings.
\begin{figure}[!t]
\centering
\includegraphics[width=3.5in]{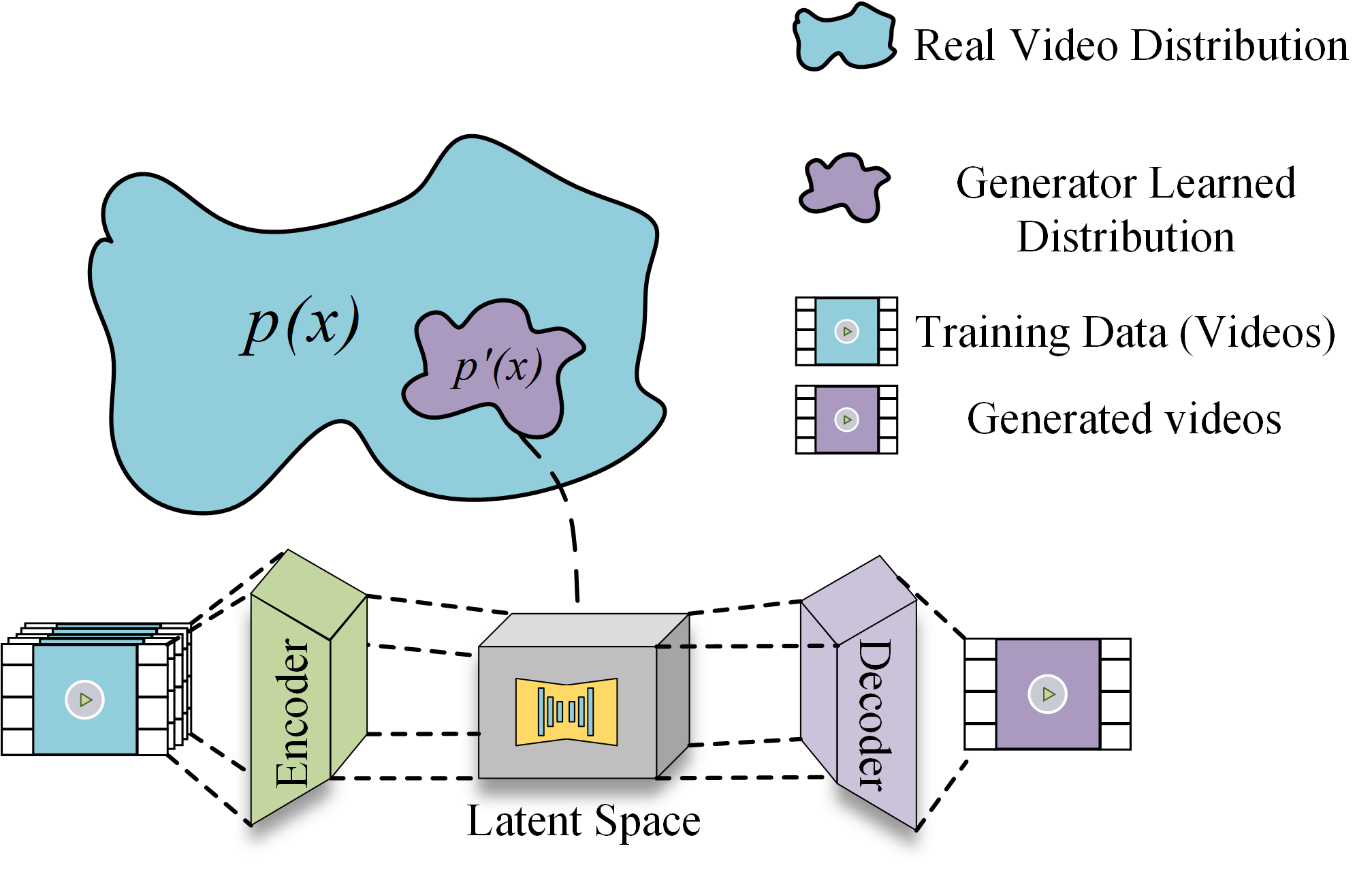}
\caption{Limitations of current generators}
\label{fig_1}
\end{figure}

To address the foregoing challenges, we introduce DBINDS (Detection Based on Initial Noise Difference Sequence), a diffusion model inversion–based approach for detecting AI‑generated videos. Unlike static image synthesis, video generation must jointly capture inter‑frame temporal dependencies while maintaining persistent spatial detail to ensure coherence across time and space. Systems such as Sora map full video sequences to a unified latent space and couple Transformer backbones with diffusion modeling, yielding marked gains in long‑horizon temporal coherence and spatial consistency relative to frame‑by‑frame methods\cite{res2}; related strategies (e.g., temporal‑dynamics modeling\cite{hong2022cogvideolargescalepretrainingtexttovideo,yuan2024mora}, optical‑flow guidance\cite{liang2024movideo}, latent‑space regularization\cite{wang2024diffperformer,wang2023videolcm,ni2024ti2v}, and temporal‑continuity losses\cite{zhang2025training}) likewise improve perceptual quality. Nevertheless, real‑world videos are governed by complex physical processes and semantic events and are further perturbed by stochastic noise from acquisition and transmission, so their latent representations often exhibit complex, high‑dimensional dynamics that no single model is likely to fully capture. Moreover, because training data are limited, the learned distribution typically covers only a subset of the true video distribution—i.e., the modeled distribution $p'(x)$ lies in a local region of $p(x)$; As illustrated in Fig. 1, after videos are encoded into latent space and then decoded, the generator synthesizes samples within the learned support, leaving substantial portions of the true distribution and its dynamics uncovered. This motivates analysis of inter‑frame transitions in latent space: if such transitions exhibit regularities, they can help discriminate real from generated videos. We therefore perform diffusion model inversion to map outputs back to intermediate or initial latent variables, establishing an approximately invertible correspondence that localizes semantically meaningful intermediate states and enables downstream analysis. Building on O‑BELM\cite{wang2024belm}, we accurately recover per‑frame initial noise (the “starting point”). By differencing the starting points of adjacent frames, we obtain a sequence of noise differences that captures inter‑frame state changes. Over this sequence, we extract and fuse multiple feature descriptors and train an efficient discriminator using cost‑sensitive LightGBM with Bayesian hyperparameter optimization. This paper makes the following contributions:
\begin{itemize}
\item{To our knowledge, we are the first to introduce diffusion model inversion into AI‑generated‑video detection and to propose latent‑space feature analysis based on the Initial Noise Difference Sequence (INDS), effecting a shift from pixel‑domain to latent‑space detection and providing a new path toward cross‑generator generalization.}
\item{We design a staged INDS feature‑analysis and optimization framework comprising multi‑dimensional feature integration, embedded feature preselection, cross‑feature interaction enhancement, and hyperparameter optimization. The pipeline balances discriminative power and generalization, reduces feature redundancy and model complexity, and comprehensively exploits the discriminative information encoded in INDS.}
\item{We conduct systematic experiments, including ablations on each feature domain and their combinations, and comparisons with mainstream detectors and vision backbones. The combination of spatiotemporal correlation features and spatiotemporal texture features achieves the best accuracy and robustness across multiple unseen generators and real‑video datasets (overall accuracy of 78.08\%), demonstrating strong zero‑shot transfer and practical utility; we also examine the performance–efficiency trade‑off under different inversion‑parameter settings.}
\end{itemize}

\section{Related Work}
\subsection{AI‑generated‑video Detection}
Prior work on detecting AI-generated content in the visual domain has primarily focused on two areas: AI-generated image detection and face-swap detection. Both areas have attracted substantial research interest and remain highly active, giving rise to multiple methodological families. For image detection, approaches include artifact-based, frequency-domain, and image-encoder–based methods. Artifact-based methods identify characteristic artifacts in generated images (e.g., CNNDetection\cite{Wang_2020_CVPR} trains deep binary classifiers). Frequency-domain methods leverage the fast Fourier transform to reveal distinctive spectral patterns\cite{frank2020leveraging}. Encoder-based methods exploit the representational spaces of pretrained models (e.g., CLIP\cite{radford2021learningtransferablevisualmodels}).The deepfake (facial‑manipulation)  detection not only targets facial-region artifacts but also leverages physiological and temporal cues, such as abnormal eye blinking\cite{8630787}, inconsistencies in rPPG-derived heart-rate signals\cite{article}, anomalous gaze-direction distributions\cite{peng2024deepfakes}, and audio–visual (speech–lip) asynchrony\cite{agarwal2020detecting}. Meanwhile, with the rapid development of general-purpose video generation (e.g., text-to-video and image-to-video), the detection of AI-generated videos is attracting increasing attention as an emerging direction, yet it remains at an early stage and presents distinct technical challenges.

Feature-extraction strategies have evolved from a primary focus on spatial artifacts to a stronger emphasis on temporal cues. AIGVDet \cite{AIGVDet} employs a dual-branch spatiotemporal CNN to model spatial cues in RGB frames and temporal cues in optical-flow fields.\cite{res3} demonstrated through probing experiments that conventional spatiotemporal CNNs tend to over-rely on spatial artifacts—effectively reducing 3D video detection to a 2D image problem—and that such spatial artifacts lack cross-generator consistency, which hinders generalization to unseen generators. In response, DeCoF\cite{res3} suppresses spatial-artifact interference via a CLIP‑ViT backbone and targets temporal artifacts that capture inter-frame consistency, markedly improving cross‑generator generalization.

Detection architectures have diversified to address heterogeneous artifact types. DeMamba \cite{res5} leverages the Detail Mamba module with a continuous scanning strategy to capture complex spatiotemporal dependencies. DuB3D \cite{ji2024distinguishfakevideosunleashing} adopts a dual‑branch design, with one branch for appearance cues and another that uses optical flow to extract motion features. For diffusion‑generated videos, MM‑Det \cite{NEURIPS2024_dccbeb7a} couples the perceptual capabilities of large multimodal models with spatiotemporal feature enhancement. Building on appearance, motion, and geometry as three basic dimensions, \cite{chang2024mattersdetectingaigeneratedvideos} propose an expert‑ensemble model that integrates multi‑dimensional information and can detect Sora‑generated videos unseen during training.

Generalization and robustness have become central concerns. Cross‑generator generalization remains limited, motivating advances in both datasets and methods. On the dataset side, the field has progressed from early GVD\cite{AIGVDet} to the million‑scale GenVideo\cite{res5} and to GenVidBench\cite{GenVidbench} with cross‑source settings, increasing both scale and diversity. On the method side, DIVID \cite{liu2024turnsimrealrobust} exploits the generality of diffusion reconstruction error to strengthen generalization, while \cite{he2024exposingaigeneratedvideosbenchmark} design a channel‑attention fusion module that combines local motion cues with global appearance changes. Nevertheless, robustness remains limited under post‑processing operations such as compression, geometric transformations, and noise perturbations.

While existing methods have achieved measurable gains on standard benchmarks, the rapid iteration of generative models continues to expose limitations in detector generalization and robustness. Notably, detecting videos from previously unseen generators under few-shot conditions remains challenging, indicating that numerous open problems persist in this area.

\subsection{Datasets for AI‑Generated Videos}
Early resources focus on deepfake‑style manipulations, such as DFDC\cite{dolhansky2020deepfakedetectionchallengedfdc}, FaceForensics++\cite{Rossler_2019_ICCV}, and Celeb‑DF\cite{Li_2020_CVPR}. Built primarily around facial tampering, these datasets have accelerated progress in face‑forgery detection but cover a relatively narrow range of scenes and semantics.

To address detection of purely generated videos, several benchmarks have appeared in recent years. Representative examples include:

{\bf{Generated Video Dataset (GVD)\cite{AIGVDet}:}}contains about 11k video samples spanning 11 generation models. However, it lacks semantic labels, source metadata, and paired samples, which hampers fine‑grained analysis.

{\bf{GenVideo\cite{res5}:}}is a large‑scale (approximately 2.271M samples) and covering multiple generators, but likewise without semantic content analysis or cross‑source splits; potential train–test content overlap can reduce task difficulty.

{\bf{Generated Video Forensics (GVF)\cite{res3}:}} provides paired videos, semantic labels, and generation prompts, but the overall size is only ~2.8k, which is insufficient for training and evaluating complex detectors.

{\bf{GenVidBench\cite{GenVidbench}:}} was proposed to mitigate these limitations and substantially upgrades the infrastructure for generated‑video detection. Its main advantages are:
\begin{itemize}
\item{Scale and quality: GenVidBench exceeds 140k videos, covering eight state‑of‑the‑art generators (e.g., Pika, MuseV, SVD) and two real‑video sources (Vript, HD‑VG‑130M), with broad variation in resolution, frame rate, and generation mode (text‑to‑video and image‑to‑video).}

\item{Cross‑source and cross‑generator splits: The benchmark adopts “cross‑source–cross‑generator” train-test partitions to reduce reliance on content‑specific or model‑specific cues, better reflecting real‑world scenarios with unknown sources and generators.}

\item{Content diversity and rich semantic annotations: Videos span a wide range of categories—for example, animals (9.2\%), buildings (23.5\%), natural scenes (20.0\%), plants (21.2\%), cartoons (12.4\%), food (10.6\%), games (10.4\%), vehicles (15.3\%), and other (5.0\%)—and are annotated along multiple dimensions (object categories, action types, scene locations, etc.), covering contexts such as humans, animals, transportation, and nature. These annotations enable fine‑grained analysis and strengthen evaluation across diverse conditions, supporting more thorough testing of detectors in complex environments.}

\item{Increased task difficulty: Empirically, several advanced models (e.g., VideoSwin, MViTv2, TimeSformer) attain substantially lower accuracy on GenVidBench than on prior datasets; for instance, SlowFast exceeds 97\% on deepfake videos but reaches only 41.66\% on GenVidBench. This indicates higher realism, complexity, and challenge, and thus greater research value.}
\end{itemize}

Overall, GenVidBench improves upon prior generated‑video datasets in scale, diversity of generation methods, breadth of semantic labels, and task design. Its combination of diverse generators and rich video content provides a solid foundation for studying the detection of AI‑generated videos and for assessing cross‑scene and cross‑model generalization. In this work, we particularly leverage its content diversity as a comprehensive testbed for evaluating our proposed approach.

\subsection{Diffusion Model Inversion}
Denoising diffusion probabilistic models (DDPMs)\cite{NEURIPS2020_4c5bcfec,pmlr-v37-sohl-dickstein15} generate data by progressively denoising a latent initialized from a Gaussian prior via a learned reverse‑time process. To support tasks such as image editing, interpolation, and reconstruction, increasing attention has turned to inversion—that is, recovering from a given sample the corresponding initial latent (e.g., the starting noise) that would reproduce it under a specified sampler. Inversion fidelity directly affects controllability and interpretability; consequently, designing stable, (approximately) reversible, and accurate inversion procedures has become a key topic in diffusion modeling.

An early deterministic approach is DDIM (Denoising Diffusion Implicit Models), which treats the reverse dynamics as an ordinary differential equation (ODE) and discretizes it for inference. In practice, however, DDIM‑based\cite{song2021denoising} inversion often exhibits path inconsistency: the recovered latent, when propagated forward with the same sampler, fails to exactly reconstruct the original sample. This lack of path closure—arising from discretization and estimation errors—is particularly apparent in reconstruction and style‑preservation tasks.

To address this structural inconsistency, bidirectional explicit linear multistep (BELM)\cite{wang2024belm} methods have been proposed as a unified numerical framework. By casting sampling as explicit linear‑multistep integration, BELM derives paired forward and reverse updates that are mutually conforming, yielding near‑invertible trajectories under the chosen discretization. Building on this framework, the optimal BELM (O‑BELM) further minimizes the local truncation error (LTE), improving bidirectional consistency and sample quality while maintaining numerical accuracy.

Within the O‑BELM framework, the forward sampling process is constructed using a variable‑step‑size, variable‑formula (VSVF) multistep method; its update equation is as follows:

\begin{equation}
x_{i-1}=\sum_{j=1}^ka_{i,j}\cdot x_{i+j}+\sum_{j=0}^{k-1}b_{i,j}\cdot h_{i+j-1}\cdot\epsilon_\theta(x_{i+j},\sigma_{i+j})
\end{equation}

where a$_{\mathrm{i,j}}$ and $b_\mathrm{i,j}$ are dynamic weighting coefficients that depend on the step size $h_\mathrm{i}$; $\epsilon_\mathrm{\theta}(x,t)$ is the noise predictor parameterized by a deep neural network, and $\sigma_{t}$ denotes the noise‑scheduling function in the diffusion process. The corresponding inversion procedure is governed by the following core inversion equation:

\begin{align}
x_{i-1} &= \frac{h_i^2}{h_{i+1}^2}\cdot\frac{\alpha_{i-1}}{\alpha_{i+1}}x_{i+1} 
+ \frac{h_{i+1}^2-h_i^2}{h_{i+1}^2}\cdot\frac{\alpha_{i-1}}{\alpha_i}x_i \notag \\
&\quad - \frac{h_i(h_i+h_{i+1})}{h_{i+1}}\cdot\alpha_{i-1}\cdot\epsilon_\theta(x_i,i)
\end{align}

Where $\alpha_{t}$ denotes the mean‑scaling coefficient at step $t$. This formulation is designed to ensure strong structural alignment between the inversion trajectory and the forward trajectory and supports dynamic step‑size scheduling to accommodate different diffusion schedules. The introduction of O‑BELM substantially improves the theoretical soundness and empirical performance of diffusion models in inversion settings. Empirical evidence indicates that O‑BELM not only enhances sample quality in unconditional image generation but also achieves stronger fidelity and detail preservation in conditional tasks such as image reconstruction and editing, thereby providing an effective route to high‑accuracy inversion sampling. 

Building on the O‑BELM framework, our DBINDS approach detects AI‑generated videos by analyzing the precisely inverted initial noise (the “starting point”).

\subsection{Efficient gradient boosting: LightGBM}
In high‑dimensional, large‑scale settings, conventional gradient‑boosted decision trees (GBDTs) \cite{friedman2001greedy} face substantial challenges in training efficiency and memory usage. LightGBM (Light Gradient Boosting Machine)\cite{ke2017lightgbm}, proposed by Microsoft, is an efficient GBDT implementation tailored for big‑data scenarios and is widely used in classification, ranking, and regression.

To alleviate the scalability bottlenecks of standard GBDT training, LightGBM introduces two key optimizations. Gradient‑based One‑Side Sampling (GOSS) retains samples with large gradients while randomly subsampling those with small gradients and compensates the latter in information‑gain computation. This enables accurate split finding from a reduced subset and markedly lowers training complexity. Exclusive Feature Bundling (EFB) merges mutually exclusive features into compact bundles, effectively reducing dimensionality and mitigating redundancy in high‑dimensional sparse data.

Empirical studies report that LightGBM achieves notable advantages over comparable implementations (e.g., XGBoost\cite{chen2016xgboost}) on multiple public datasets. In terms of training efficiency, speedups of up to an order of magnitude (reported up to 20×) have been observed. Moreover, even with instance sampling and feature bundling, classification and ranking performance remains close to that of full‑data training, supporting the effectiveness of these optimizations. LightGBM also exploits feature sparsity to reduce memory footprint and supports efficient parallel training.

Given these strengths in handling large sample sizes with sparse, high‑dimensional features, LightGBM is well suited to AI‑generated video detection, where heterogeneous feature families must be fused. In this work, we adopt a cost‑sensitive LightGBM as the core classifier and apply Gaussian‑process–based Bayesian hyperparameter optimization \cite{snoek2012practical} to automatically tune the configuration, further improving performance on video authenticity classification.

\section{Proposed Method}

This section elaborates the core ideas and implementation of DBINDS. DBINDS follows a latent‑space detection strategy: it applies diffusion‑model inversion to each video frame to recover the per‑frame initial noise, constructs an Initial Noise Difference Sequence (INDS) by differencing adjacent frames, and conducts multi‑dimensional feature analysis on this sequence to reveal systematic discrepancies between AI‑generated and real videos in latent space. Compared with conventional vision methods that operate directly on pixel‑domain content, this inversion‑driven analysis yields improved detection performance in our experiments and affords clearer interpretability, as the features are grounded in the generative trajectory rather than visible artifacts. 

\begin{figure*}[!t]
\centering
\includegraphics[width=\textwidth]{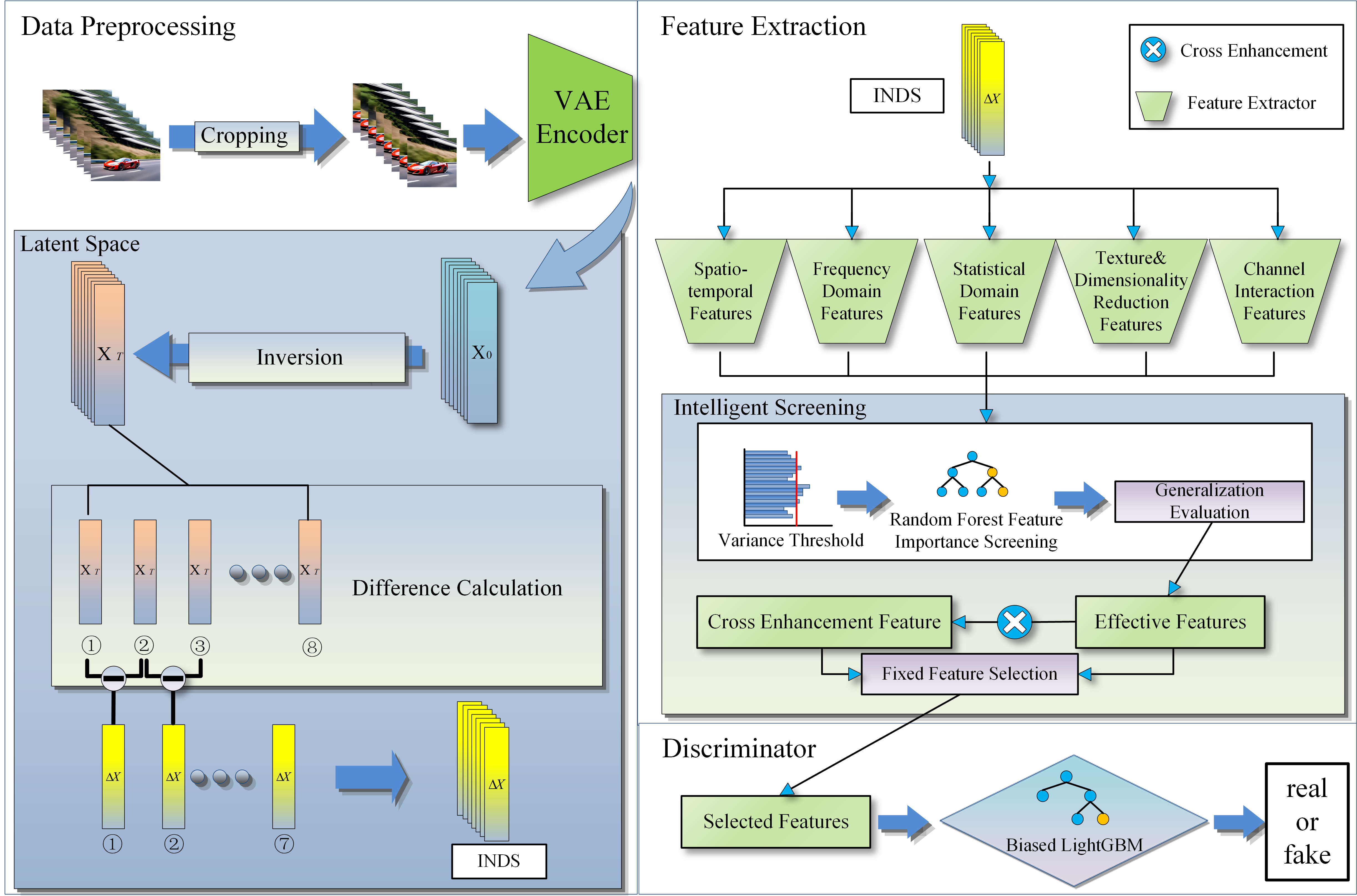}
\caption{INDS feature‑analysis pipeline in DBINDS. The pipeline first performs data preprocessing (cropping) and VAE encoding. In latent space, O‑BELM–based inversion is applied and difference calculation is conducted to derive the Initial Noise Difference Sequence (INDS). From INDS, features are extracted in five domains in parallel: spatiotemporal features, frequency‑domain features, statistical‑domain features, texture‑and‑dimensionality‑reduction features, and channel‑interaction features. An intelligent screening module then applies variance‑threshold filtering and random‑forest–based importance screening, with a generalization‑evaluation loop; the retained features are further combined via a fixed feature‑selection step and a cross‑enhancement module to form more discriminative representations. Finally, the selected features are fed into a cost‑sensitive (biased) LightGBM classifier to decide whether the video is real or AI‑generated.}
\label{fig_2}
\end{figure*}

\subsection{Overall architecture}

As illustrated in Fig. 2, DBINDS comprises four stages: construction of the Initial Noise Difference Sequence (INDS), multi‑domain feature extraction, automated feature selection with cross‑feature interaction enhancement, and classification. Given an input video $V = {I}_{t=1}^T$, where $I_t$ denotes the $t$‑th frame and $T$ is the sequence length, we first recover per‑frame initial noise via O‑BELM‑based inversion and build the INDS by differencing adjacent frames. The resulting sequence is then processed by the feature pipeline to derive discriminative descriptors, which are fed to the classifier to separate real and AI‑generated videos.

\subsection{INDS construction}

To ensure computational efficiency and feature consistency, we adopt a standardized preprocessing routine. From each video, we uniformly sample eight frames at fixed intervals (default stride of 2; for short clips the interval is adapted), thereby increasing visual change between adjacent samples and strengthening the discriminativeness of subsequent difference features. To remove dimensional mismatches in the inverted noise caused by varying input resolutions, all frames are standardized to 512×512: larger images are center‑cropped to preserve core content, while smaller ones are zero‑padded at the borders. The frames are then encoded with a VAE and inverted using O-BELM to recover the initial noise per frame. The INDS is then constructed by differencing these adjacent noise maps. This standardization preserves comparability across videos while retaining key information in the central region, providing a stable basis for latent‑space analysis.

Building on the O‑BELM framework outlined in Section~{\Large \textsc{ii}}, we perform diffusion‑model inversion on each preprocessed frame $I_t$. Empirical results reported in the original work indicate that a 10‑step schedule suffices to bring the reconstruction mean‑squared error (MSE) to a stable plateau while achieving high‑fidelity inversion. Accordingly, we adopt a 10‑step configuration to balance accuracy and computational cost. The O‑BELM inversion yields the initial noise for each frame $\varepsilon_{\mathrm{t}}\in\mathbb{R}^{4\times64\times64}$, which we then arrange in temporal order to form the initial‑noise sequence:

\begin{equation}
\mathcal{E}=\{\varepsilon_1,\varepsilon_2,\ldots,\varepsilon_8\}
\end{equation}

Based on the initial‑noise sequence obtained via inversion, we construct the difference sequence between consecutive frames:

\begin{equation}
\mathcal{D}=\{d_t=\varepsilon_{t+1}-\varepsilon_t|t=1,2,\ldots,7\}
\end{equation}

Each $d_t\in\mathbb{R}^{4\times64\times64}$ encodes the spatiotemporal variation between neighboring frames in latent space. This difference representation is motivated by a working hypothesis: real and AI‑generated videos follow distinct temporal evolution patterns in latent space. In particular, videos produced by generative models are constrained by their internal algorithms and thus exhibit characteristic dynamics in the noise space, distinguishable from the natural variability observed in real videos. 

A further advantage of differencing is information refinement. The raw initial noise contains substantial, and inevitably redundant, information. By differencing adjacent frames, salient changes are emphasized while redundancy is reduced, enabling more precise extraction and retention of task‑relevant spatiotemporal dynamics. This improves computational efficiency and increases sensitivity to subtle manipulation traces. 

\subsection{Integrated feature‑analysis module}
Given that classifiers built on a single feature domain tend to be constrained by specific data distributions when analyzing the noise‑difference sequence, thereby limiting cross‑dataset generalization, we fuse features from multiple domains to obtain complementary evidence and improve overall performance. Accordingly, we design a comprehensive feature‑analysis framework spanning spatiotemporal, frequency‑domain, and statistical features. Although these techniques originate from conventional image analysis, the initial‑noise difference sequence (INDS) is shape‑compatible with images; thus, they can be repurposed for the latent‑noise space, enabling a shift in the detection paradigm from the pixel domain to the latent domain. Moreover, within the feature-engineering workflow, we employ two completion strategies to maintain a consistent and robust feature space. Zero padding extends representations to a predefined dimensionality when the number of principal components or the available texture statistics is insufficient. Value imputation is applied in exceptional cases such as operator failure or missing statistics, substituting the median or zero to prevent invalid entries in the feature vector. These procedures serve solely to safeguard the structural integrity of the feature matrix and do not introduce additional discriminative information.

\subsubsection{Spatiotemporal‑domain feature analysis}

{\bf{Spatiotemporal Energy Distribution Characteristics:}} The analysis of energy distribution is based on signal processing theory, where the strength of the signal at each spatiotemporal point is measured by its intensity. To extend this method further, we apply it to the spatiotemporal analysis of noise signal distributions, calculating the energy sequences in both the time and space dimensions. By doing so, we capture the energy variation patterns introduced in space due to the signal generation process. During the generation process, due to the inherent characteristics of the signal, the energy distribution may exhibit patterns that differ from those of real videos. Through this analysis, we are able to uncover the feature variations of different signal sources across spatiotemporal dimensions, thus providing robust support for video content recognition.

The local spatiotemporal energy combines all time steps, considering both the spatial energy at each position and the overall anomaly degree reflected in the spatiotemporal dimensions. The energy sequences are then obtained for both temporal and spatial energy variation patterns. The equations are as follows:

\begin{equation}
E_{global}=\sum_{t=1}^7\sum_{c=1}^4\sum_{h=1}^{64}\sum_{w=1}^{64}d_t^2(c,h,w),
\end{equation}
\begin{equation}
E_{temporal}(t)=\sum_{c,h,w}d_t^2(c,h,w),
\end{equation}
\begin{equation}
\quad E_{spatial}(h,w)=\sum_{t,c}d_t^2(c,h,w)
\end{equation}

{\bf{Spatiotemporal Gradient Fusion Features: }} Gradient analysis is a fundamental technique in image processing, conventionally employed for detecting edges and textural details. We apply this analysis to the noise difference sequence to identify anomalous patterns symptomatic of the underlying generative process. The rationale is that gradient distributions in the noise residuals of authentic videos are typically stochastic and unstructured. Conversely, those from generated videos often manifest discernible structural patterns—artifacts resulting from the inherent constraints of the synthesis model. To this end, we fuse temporal and spatial gradient information to construct a unified three-dimensional gradient magnitude, denoted as $G_{\text{magnitude}}$. This magnitude is derived from the temporal gradient ($G_{\text{temporal}}$) and the spatial gradients along the height ($G_{\text{spatial}}^{(h)}$) and width ($G_{\text{spatial}}^{(w)}$) dimensions. All components are functions of time ($t$), height ($h$), and width ($w$). The relationship is formulated as follows:

\begin{align}
\mathrm{G_{magnitude}} = \sqrt{\mathrm{G_{temporal}^2} + \mathrm{G_{spatial}^{(h)2}} + \mathrm{G_{spatial}^{(w)2}}}
\end{align}

{\bf{Spatiotemporal correlation features: }} Correlation analysis is widely used to quantify linear dependence, and we additionally compute mutual information to capture nonlinear dependence. We adapt these tools to the initial noise‑difference sequence (INDS) to probe regularities induced by generative modeling in latent space. Concretely, we assess inter‑frame correlations via Pearson’s r to quantify temporal consistency, and per‑location temporal correlations by examining the correlation of each spatial location’s trajectory; we also compute temporal and spatial autocorrelations to characterize dependencies along the time axis and within each frame. In addition, mutual information between the first and last frames is used to capture nonlinear dependence, and cross spatiotemporal correlations relate global, frame‑level descriptors to local, patch‑level features. Consistent with our hypothesis, AI-generated videos, being constrained by algorithmic and training regimes, tend to exhibit more regular, model-dependent correlation structures across neighboring frames and spatial locations, whereas real videos display more natural and diverse patterns. For each correlation map or series, we summarize the results using mean, standard deviation, variance, maximum, minimum, median, skewness, and kurtosis, which are then passed to the subsequent selection and fusion stages.

\subsubsection{Frequency‑domain features}
{\bf{Depth-Spectrum Analysis:}} is a fundamental tool in signal processing and is widely used in imaging to characterize periodic patterns and frequency content. We extend the fast Fourier transform (FFT) from the pixel domain to the latent‑noise domain of the initial noise‑difference sequence (INDS). Our working hypothesis is that the frequency preferences learned during training and manifested during inference leave identifiable spectral “fingerprints” in latent space; owing to architectural and training differences, such spectra for AI‑generated videos exhibit systematic deviations from those of real videos. Concretely, we apply a 1D FFT to the temporal trajectory at each representative spatial location and a 2D FFT to selected difference frames to obtain spatial spectra. To balance efficiency and coverage, five spatial sites (the four corners and the center) and five uniformly spaced time points (start, 1/4, middle, 3/4, end) are used. Frequency‑distribution features are then derived via radial averaging. By computing the radial frequency distribution, we derive frequency‑distribution features:

\begin{equation}
P_{radial}(r)=\frac{1}{N_r}\sum_{u^2+v^2=r^2}\left|\mathcal{F}_{spatial}(t,u,v)\right|^2
\end{equation}

where $r$ denotes the radial distance from the spectral origin, $N_r$ is the number of coefficients on the corresponding annulus, and $F_spatial$ denotes the 2‑D Fourier spectrum. This analysis reveals generator‑specific patterns introduced in the frequency domain. We partition the spectrum into low‑, mid‑, and high‑frequency bands and compute the energy proportion within each band; in addition, we derive band‑ratio features (low/mid/high), perform spectral‑peak detection, and measure spatiotemporal spectral consistency to capture regularities in the frequency‑modeling behavior of generative models.

{\bf{Wavelet-based multiscale analysis:}}  We apply discrete wavelet transforms (DWT) to the initial noise difference sequence (INDS) for multiresolution characterization. Using standard bases—Daubechies‑4 (db4), Haar, and biorthogonal 2.2 (bior2.2)—we perform two-level decompositions by default. Along the temporal pathway, representative spatial sites are sampled on a 16‑pixel grid; the pixel-level temporal trajectory at each site is processed with a 1D DWT. Along the spatial pathway, time is subsampled with a stride of two frames; the corresponding noise‑difference frames are each processed independently with a 2D DWT (per‑frame, rather than jointly over all seven differences). From the approximation and detail subbands at each level, we extract statistical descriptors, and we construct fused descriptors and correlation measures across temporal and spatial wavelet coefficients to expose discriminative cues in latent space across scales. This multiscale representation helps capture systematic differences between AI‑generated and real videos in their signal decompositions.

\subsubsection{Statistical-domain feature modeling}
{\bf{Higher-order statistical features:}} First- and second-order statistics capture only basic aspects of a distribution. To more fully characterize the initial noise difference sequence (INDS), we adopt higher-order moment analysis: third–sixth moments are computed to quantify skewness, kurtosis, and higher-order shape attributes. We additionally employ L‑moments—based on order statistics—as a robust complement with improved resistance to outliers. These descriptors are extracted at global, temporal, and spatial scales to reveal potential systematic deviations in the distributional modeling of generative processes. In practice, due to constraints from model capacity and training data, generative models often struggle to reproduce the complex statistical structure of real data, and such differences are more apparent in higher-order moments. 

{\bf{Local variability analysis:}} An 8×8 sliding window is used to compute local statistics on the INDS, including within-window variance, entropy, and standard deviation along the temporal axis, thereby capturing local intensity and temporal fluctuation. The underlying hypothesis is that effective receptive fields and parametric constraints in generators induce local variability patterns that differ from those of real data. In addition, Sobel edge maps are computed at each time step, and the temporal correlation of edge magnitudes between adjacent frames is measured to assess edge preservation and change dynamics in generated videos. 

\subsubsection{Texture and dimensionality‑reduction features}
{\bf{Texture features:}} We extract texture descriptors on key INDS frames (first, middle, last) using the gray‑level co‑occurrence matrix (GLCM). GLCMs are computed with offset distance 1 at orientations 0° and 90°, and standard statistics (contrast, dissimilarity, homogeneity, energy) are derived; gray‑level quantization and preprocessing details are documented in the appendix. Since generative models often exhibit characteristic preferences and constraints in texture synthesis, texture patterns in the noise domain may differ systematically from those of real videos. To assess spatiotemporal texture consistency, we also compute frame‑to‑frame Pearson correlation and structural similarity (SSIM).

{\bf{PCA features:}} The INDS is reshaped into a 2‑D matrix whose rows index time steps and whose columns are the flattened spatial‑and‑channel dimension (D=C×H×W). After z‑score normalization, principal component analysis is performed and the top seven components are retained. For each component’s score vector, we compute summary statistics—mean, standard deviation, variance, maximum, minimum, median, skewness, kurtosis, energy, and L1 norm—and record the explained‑variance ratio. PCA helps identify dominant directions of variation in the noise space: real videos, due to stronger natural randomness, tend to yield more dispersed component contributions, whereas model‑generated videos, constrained by parameterization and training distributions, often concentrate variation on a few components.

{\bf{LBP features:}} Local binary patterns (LBP) with radius r=1 and P=8 neighbors are computed on the key frames, and histogram‑based statistics are extracted (optionally using the uniform‑LBP variant). LBP captures local texture regularities; generators may produce repetitive or over‑regular local patterns, which are preserved in the noise domain and can be exploited for detection.
Overall, these descriptors quantify differences between model‑generated and real videos in spatial structure and dominant variation modes within the latent noise space, offering complementary evidence given the architectural and data constraints of generative models.

\subsection{Automated feature optimization}
Following the multi‑domain extraction, we obtain a candidate pool with several thousand dimensions. This pool includes both effective features produced by the analysis modules and padding entries introduced to enforce a fixed vector length. Both types are subjected to a unified importance assessment so that only dimensions with material contribution to classification are retained. Given the task complexity and the high dimensionality of the feature space, we adopt an experiment‑driven, modular optimization strategy that combines pragmatic screening with parameter tuning to progressively identify an effective subset.

In the first stage, the raw candidate space contains substantial redundancy and noise; using it as is would exacerbate the curse of dimensionality and lead to overfitting. We therefore employ a two‑step coarse screening procedure grounded in statistical properties and learning‑based estimates to reduce dimensionality while preserving salient information. The process begins with variance‑threshold filtering, which removes near‑constant features with limited discriminative power; the precise criterion is specified in the corresponding equation.

\begin{equation}
\mathcal{F}_{valid}=\{f_i|\mathrm{Var}(f_i)>\sigma_{min}\}
\end{equation}

where $\sigma_{min}$ denotes the preset variance threshold. This step removes constant and near‑constant features, providing a clean basis for subsequent analysis. We then estimate feature importance with a random forest and rank features by the mean decrease in impurity (Gini importance), retaining those with higher scores. After these two coarse‑filtering steps, low‑quality and ineffective features are largely eliminated. To further capture nonlinear relationships among features, we introduce a cross‑feature enhancement strategy to derive more discriminative representations. The strategy includes product crosses, ratio crosses, and composite crosses; the corresponding formulas are given below.

\begin{align}
f_{cross}^{(1)}=f_i\odot f_j,\quad f_{cross}^{(2)}=\frac{f_i}{f_j+\epsilon},\quad f_{cross}^{(3)}=\alpha f_i+\beta f_j
\end{align}

Among them, $\odot$ denotes the Hadamard product, $\epsilon$ is a small constant to prevent division by zero, and $\alpha$ and $\beta$ are the weighting coefficients for the combination.

After obtaining the strengthened features via feature augmentation, we subsequently evaluated the generalization capability of all features. Feature selection was ultimately determined by a composite metric that integrates both the discriminative power for the training data and the generalizability to validation data. Specifically, we combined the F-statistic (discriminative power on the training set, \(S_{\text{train}}\)) and the absolute value of the regression coefficient on the validation set (\(S_{\text{val}}\), representing the importance on validation data), as formulated below:

\begin{equation}
\S_{\text{combined}} = 0.4 \cdot S_{\text{train}} + 0.6 \cdot S_{\text{val}}\
\end{equation}

Here, the weighting scheme manifests the emphasis laid on the generalizability of selected features, aiming to ensure robust model performance on unseen data. By this streamlined selection procedure, candidate features are iteratively narrowed down, significantly reducing feature dimensionality. This not only preserves the essential discriminative information but also substantially mitigates feature redundancy. The approach lays a solid foundation for subsequent target subset determination.

Building on the pre‑screened features, we conduct an experiment‑driven exploration to identify an optimal subset. The process emphasizes systematic evaluation of alternative feature combinations rather than reliance on a single theory‑ or heuristic‑driven filter. On this basis, we assess the following combination strategies:

\begin{itemize}
\item{{\bf{Single‑module combinations:}} use features from one module at a time to obtain its standalone discriminative power and a baseline ranking.}
\item{{\bf{Importance-ranking combinations:}} form Top‑K subsets at multiple thresholds based on random‑forest importance scores and evaluate the alignment between importance and downstream contribution.}
\item{{\bf{High-performance module combinations:}} combine top‑ranked modules in pairs, triplets, and quartets to probe synergistic gains.}
\item{{\bf{Strategy‑driven combinations:}}  design complementary schemes (e.g., energy+gradient, PCA+correlation, texture+local variability) to integrate first‑order energy cues, second‑order texture descriptors, and low‑dimensional PCA information into composite representations.}
\item{{\bf{All-module combination:}} use all screened features jointly as a reference baseline.}
\item{{\bf{Fuzzy‑matching combinations:}} To account for possible complementarities among feature dimensions in latent space, we employ a keyword‑driven fuzzy‑matching process that programmatically gathers and assembles candidate subsets. The process anchors on predefined core identifiers (PCA, energy, correlation) and allows minor spillover from similarly named features to exploit cross‑domain complementarity. After iterative tuning, the resulting subset performs better than competing combinations on the validation set and combines core module features with a small set of gradient‑, spectral‑, and texture‑based descriptors.}
\end{itemize}

\subsection{Classifier optimization and thresholding}
\begin{figure*}[!t]
\centering
\subfloat[]{\includegraphics[width=3.5in]{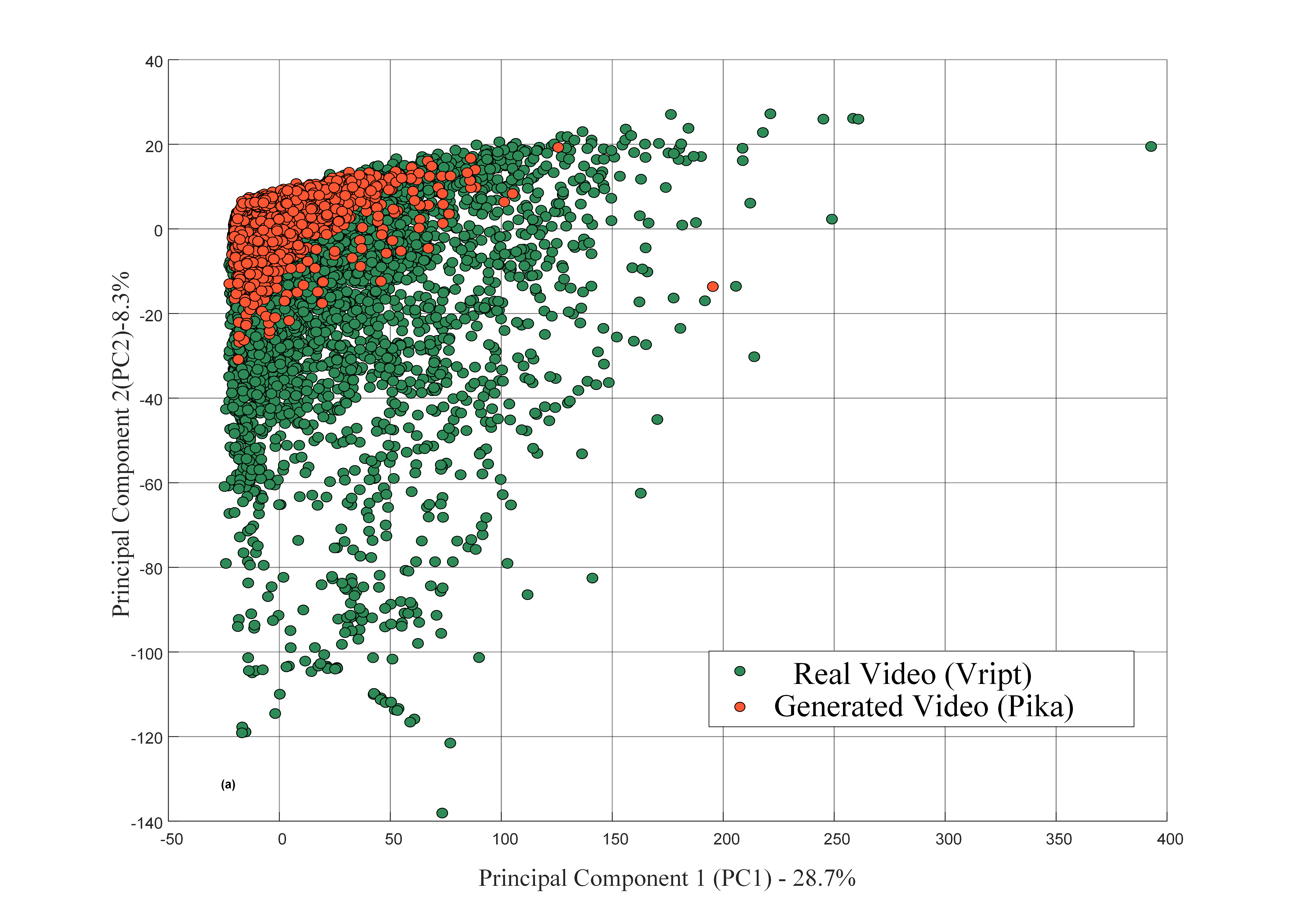}%
\label{fig_first_case}}
\hfil
\subfloat[]{\includegraphics[width=3.5in]{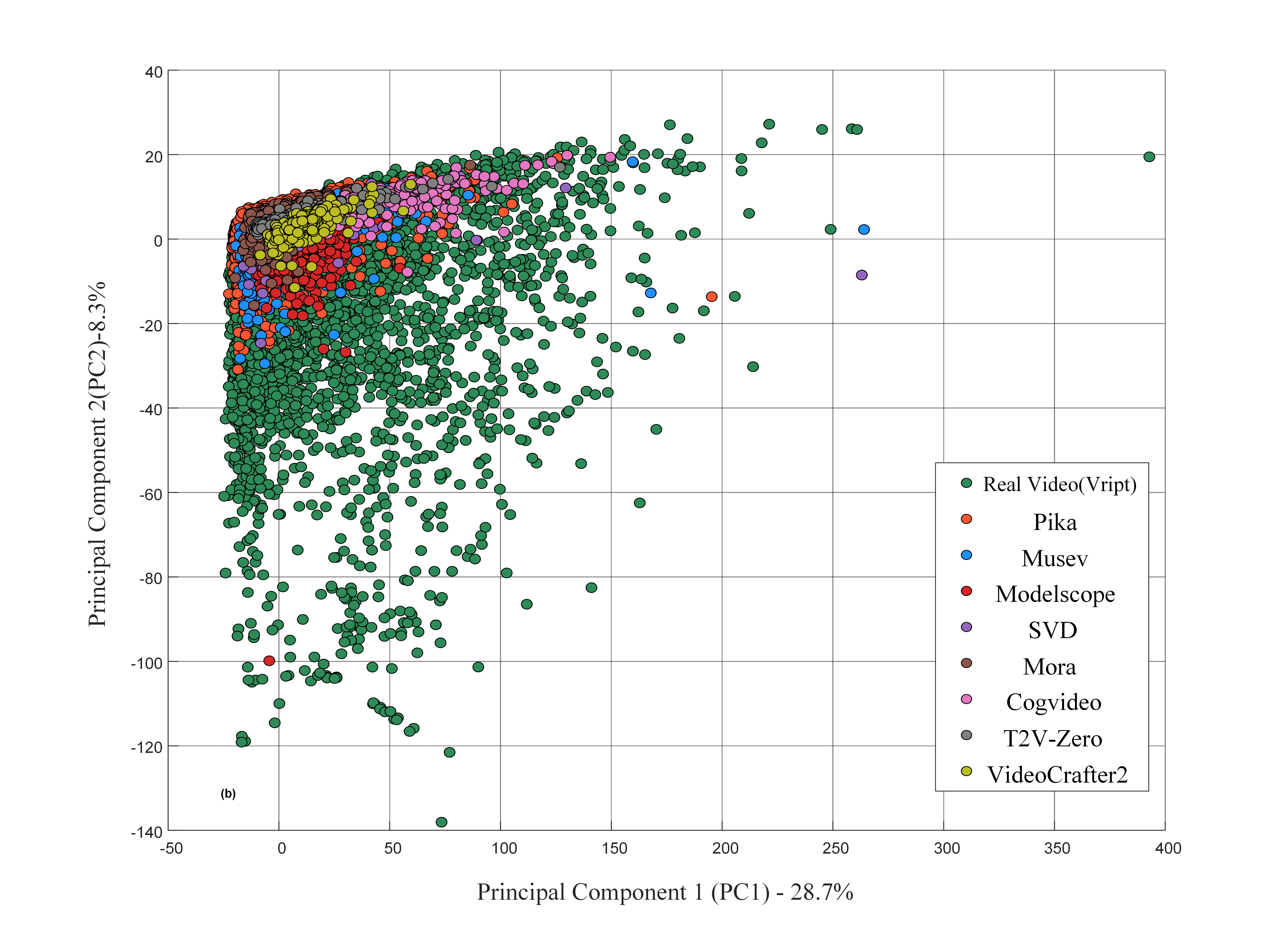}%
\label{fig_second_case}}
\caption{The PCA visualization distribution under the full feature combination. (a) PCA of training scenarios (Vript vs. Pika). (b) Feature distributions across multiple generators.}
\label{fig_sim}
\end{figure*}

Based on the finalized feature set, we optimize the LightGBM classifier within a cost-sensitive learning framework. Fig. 3 presents a PCA projection (first two principal components) computed from the concatenation of all screened features without domain grouping. It should be noted that this visualization is not derived from the optimal feature combination but rather from the aggregation of all effective features, intended to illustrate the overall distributional tendency. The visualization shows that the generator used for training (Pika) and the other seven generators occupy a consistent region near the leading edge of the “fan-shaped” distribution formed by real samples, whereas real videos spread over a broader area. This observation is consistent with our hypothesis: constraints imposed by inter-frame modeling techniques and by the scale/diversity of training data restrict the latent dynamics of generators, yielding more regular inter-frame evolution patterns for AI-generated videos. Motivated by this, we apply a slight bias in class weighting. This adjustment increases sensitivity to generated data while maintaining precision for real samples, thereby improving the balance between overall accuracy and detection recall. The rationale for this design is threefold. First, a controlled reduction in real-class accuracy can be tolerated in exchange for a higher recall of generated instances, with fine-tuning ensuring that overall accuracy is preserved. Second, it accounts for the asymmetric costs of errors in AI-content detection, where false negatives are generally more critical than false positives. Finally, the mild weighting enhances robustness to future distributional shifts, as synthetic content is expected to become increasingly similar to real data.
\begin{algorithm}[ht]
	\caption{TPE-based Bayesian hyperparameter optimization with ROC-based threshold selection}
	\label{alg:tpe_roc}
	\begin{algorithmic}[1]
		\STATE \textbf{Input:} Training/validation split $(X_{\text{train}},y_{\text{train}})$, $(X_{\text{val}},y_{\text{val}})$; LightGBM search space $\mathcal{S}$; target generated detection rate (GDR) $\tau$; positive-class weight multiplier $m$; number of trials $T$
		\STATE \textbf{Output:} Best hyperparameters $\lambda^*$, selected threshold $\theta^*$, best objective $J^*$
		
		\STATE Initialize TPE over $\mathcal{S}$; set $J^*\gets-\infty$.
		\FOR{$t=1,\dots,T$}
		\STATE Sample hyperparameters $\lambda_t$ via TPE.
		\STATE Compute balanced class weights on $y_{\text{train}}$ and multiply the generated-class weight by $m$; train a class-weighted LightGBM with $\lambda_t$ on $(X_{\text{train}},y_{\text{train}})$.
		\STATE On $X_{\text{val}}$, compute predicted probabilities $p$ and the ROC curve $(\mathbf{fpr},\mathbf{tpr},\boldsymbol{\theta}_{\text{cand}})$.
		\STATE Threshold selection:
		\STATE \hspace{1em} If there exist indices with $\mathbf{tpr}_i\geq\tau$, choose $\theta_t$ as the first threshold along the ROC traversal that achieves $\text{GDR}(\theta_t)\geq\tau$.
		\STATE \hspace{1em} Otherwise, set $\theta_t\gets\arg\max_{\theta\in\boldsymbol{\theta}_{\text{cand}}}\text{GDR}(\theta)$.
		\STATE Form binary predictions on $X_{\text{val}}$ by thresholding $p$ at $\theta_t$; compute $\text{Acc}(\theta_t)$ and $\text{GDR}(\theta_t)$.
		\IF{$\text{GDR}(\theta_t)<\tau$}
		\STATE $J_t\gets 0$
		\ELSE
		\STATE $J_t\gets 0.7\times\text{Acc}(\theta_t)+0.3\times\text{GDR}(\theta_t)$
		\ENDIF
		\STATE Update TPE with $(\lambda_t,J_t)$.
		\IF{$J_t>J^*$}
		\STATE $(\lambda^*,\theta^*,J^*)\gets(\lambda_t,\theta_t,J_t)$
		\ENDIF
		\ENDFOR
		\STATE \textbf{Return:} $(\lambda^*,\theta^*,J^*)$
		\STATE \textit{Note:} GDR denotes recall on the generated (positive) class. In tables, DetectionRate refers to GDR.
	\end{algorithmic}
\end{algorithm}
Building on this design, we further refine the model through Bayesian hyperparameter optimization using the Tree-structured Parzen Estimator (TPE), in combination with ROC-based threshold selection. In each trial, LightGBM is trained under the biased weighting scheme, and the decision threshold is determined from the validation ROC curve such that the recall requirement for generated content is satisfied whenever possible. Trials that fail to meet the constraint are discarded, while the remaining ones are evaluated using a weighted objective that balances accuracy and detection rate. The detailed procedure is summarized in Algorithm 1.

Capitalizing on staged feature optimization and hyperparameter tuning, we develop a detection framework for selecting high‑quality representations. The framework balances feature diversity with computational cost and replaces single‑rule filtering with a systematic, experiment‑driven search, enabling comprehensive and effective analysis of the INDS sequence. Compared with end‑to‑end deep classifiers, our feature‑centric paradigm computes the vast majority of descriptors via deterministic or weakly supervised signal‑processing and statistical methods, using a data‑efficient LightGBM classifier only at the final stage. This approach reduces reliance on large labeled datasets for representation learning and, at the same time, offers superior deployability in terms of computational footprint and latency.

\section{Experiments}
We evaluate our approach on the GenVidBench dataset. Unless otherwise noted, all experiments use the Bayesian‑optimized (TPE) configuration; the only exceptions are the feature‑combination ablations/comparisons and the inversion‑accuracy sensitivity study. Concretely, we first identify the best feature combination via ablations, and then apply Bayesian optimization to the LightGBM hyperparameters for that combination, which yields stronger performance than the ablation‑stage baselines on both validation and test sets. 

Unlike the pair1/pair2 multi‑source training protocol in the original GenVidBench paper, we adopt a one‑to‑many training setup: the training set contains only Pika (generated) and Vript (real) videos, with 13,850 and 20,131 samples respectively. We do not enforce class rebalancing, both to keep training conditions uniform and to reflect the practical prior that real videos are typically more abundant than generated ones. Throughout training, the model is not exposed to any other generators or real sources used at test time. This protocol is stricter than many‑to‑many cross‑source training and provides a more rigorous assessment of cross‑generator generalization. 

For testing, we construct a set of 3,000 videos by randomly sampling from the remaining eight subsets: 1,500 real videos from HD‑VG‑130M and 1,500 generated videos from seven unseen generators (approximately 214–215 per generator). This design ensures no overlap in generator types between training and testing, aligning with realistic AIGC‑detection scenarios in which generative mechanisms evolve over time and real‑world content is diverse. Except for the comparison with mainstream methods, all feature‑combination experiments, robustness evaluations, and inversion‑accuracy sensitivity tests report results on this unified test set. 

\subsection{Experimental Settings}
\subsubsection{Feature Combination Ablation and Comparative Study}
This experiment is designed to systematically evaluate the impact of different INDS feature subsets and their various combinations on detection performance, thereby exploring the relationship between feature diversity and optimal discriminative configurations. The study first examines the classification performance of individual primary feature modules, including spatiotemporal correlation, spatiotemporal texture, depth-spectrum analysis, energy distribution, and principal component analysis. We then constructed multiple feature combinations, including module-level integrations and importance-based Top-K selections. We also employed heuristic screening and keyword-based fuzzy matching to explore alternative fusion strategies. Through a series of ablation and comparative analyses, the optimal feature combination is identified using a unified test set, and its discriminative power and generalization capability are validated. It should be noted that this stage only establishes the baseline performance of feature combinations, without parameter optimization; hence the reported performance does not represent the best achievable results.

\subsubsection{Comparative Experiments}
To ensure representative coverage across research paradigms, a set of established models is selected for comparison. ResNet50 is adopted as the baseline convolutional neural network; ResNet50 with optical flow + RGB dual modalities and Demamba-XCLIP represent open-source AIGC-oriented video detection approaches. Furthermore, XCLIP\cite{xclip}, VideoMAE\cite{videomae}, and MViT V2\cite{mvit_v2} are included as mainstream large-scale pre-trained Transformer architectures. In particular, MViT V2 (Multiscale Vision Transformer V2), noted for its multi-scale information modeling and superior performance, serves as a strong representative of advanced video analysis frameworks. For fair comparison, XCLIP, VideoMAE, and MViT V2 are initialized with official pre-trained weights, with only the classification head fine-tuned on the few-shot training set. To adapt Demamba-XCLIP to the small-sample setting, its backbone parameters are largely frozen, while lightweight architectural adjustments and hyperparameter reductions are applied to decrease model complexity and improve suitability for low-resource scenarios. Training is limited to the classification head and Mamba layers. The training dataset consists of 13,850 Pika-generated samples and 20,131 Vript real samples, which constitutes a relatively small scale for video classification tasks. This setup enables rigorous evaluation under resource-constrained conditions. All models are tested on the HD-VG-130M benchmark provided by GenVidBench, covering seven additional generators and real datasets. Detection accuracy on each dataset is reported in the main results table, facilitating comprehensive assessment of generalization performance on unseen data.

\subsubsection{Robustness Evaluation}
This experiment examines the robustness of the optimal feature combinations obtained in the previous analyses. Specifically, it evaluates detection accuracy variations when tested on previously unseen generators and real datasets, and further investigates sensitivity to common post-processing perturbations such as video compression, noise injection, and resolution changes. The objective is to uncover robustness differences among feature combinations and to quantify the stability of generalization performance under diverse real-world interference scenarios.

\subsubsection{Sensitivity to Inversion Precision}
To examine the effect of diffusion model inversion parameters on detection performance and computational efficiency, this experiment compares the accuracy and computational cost of feature extraction under different inversion steps (e.g.,1, 5, 10, 15, 20). Additionally, the impact of sampling uncertainty and noise perturbations on detection accuracy is evaluated. The findings aim to inform practical parameter settings and resource allocation strategies for deployment, while quantifying the contribution of inversion strategies to both model discriminability and engineering efficiency.

\subsection{Experimental Results and Analysis}
\subsubsection{Results of Feature Combination Ablation and Comparison}

\begin{table*}[htbp]
	\centering
	\footnotesize                  
	\setlength{\tabcolsep}{2.8pt} 
	\begin{threeparttable}
		\caption{Performance of Different Feature Modules and Combinations (\%)}
		\begin{tabular*}{\textwidth}{@{\extracolsep{\fill}}l*{3}{c}@{}}
			\toprule
			Category \& Module/Combination & Overall Accuracy & Generated Detection Rate & Real-Sample Detection Rate \\
			\midrule
			\multicolumn{4}{l}{\textit{Single-Module Features}}\\
			\quad Spatiotemporal Correlation & 72.7 & 80.1 & 65.3\\
			\quad Spatiotemporal Texture & 71.2 & 80.1 & 62.3\\
			\quad Depth-Spectrum Analysis & 65.7 & 80.2 & 51.1\\
			\quad Spatiotemporal Energy Distribution & 64.2 & 80.1 & 48.3\\
			\quad Principal Component Analysis & 63.5 & 80.1 & 46.8\\
			\quad Spatiotemporal Local Variability & 61.5 & 80.3 & 42.8\\
			\quad Spatiotemporal Gradient Fusion & 61.0 & 80.3 & 41.7\\
			\quad Higher-Order Statistics & 58.0 & 80.1 & 36.0\\
			\quad Cross-Enhanced & 57.5 & 80.1 & 34.9\\
			\quad Wavelet-Based Multiscale Analysis & 55.3 & 80.1 & 30.4\\
			\midrule
			\multicolumn{4}{l}{\textit{Importance-Ranking Combinations}}\\
			\quad Top-80 Features & 74.6 & 80.1 & 69.1\\
			\quad Top-100 Features & 74.6 & 80.1 & 69.1\\
			\quad Top-400 Features & 73.9 & 80.3 & 67.5\\
			\quad Top-500 Features & 73.5 & 80.1 & 66.9\\
			\quad Top-200 Features & 73.4 & 80.3 & 66.5\\
			\quad Top-424 Features & 73.3 & 80.1 & 66.5\\
			\quad Top-20 Features & 71.6 & 80.1 & 63.2\\
			\midrule
			\multicolumn{4}{l}{\textit{High-Performance Module Combinations}}\\
			\quad Best Dual Combination & 76.3 & 80.3 & 72.3\\
			\quad Best Triple Combination & 76.1 & 80.1 & 72.1\\
			\quad Best Quadruple Combination & 73.1 & 80.1 & 66.1\\
			\midrule
			\multicolumn{4}{l}{\textit{Strategy-Driven Combinations}}\\
			\quad PCA + Correlation & 72.9 & 80.1 & 65.7\\
			\quad Statistical Features & 73.4 & 80.1 & 66.8\\
			\quad Multidimensional Spatiotemporal Features & 70.7 & 80.1 & 61.3\\
			\quad Extended Spatiotemporal Features & 71.5 & 80.1 & 62.9\\
			\quad Texture + Local Variation & 69.7 & 80.1 & 59.3\\
			\quad Energy + Gradient Fusion & 61.8 & 80.1 & 43.5\\
			\quad Cross-Feature Combination & 60.9 & 80.1 & 41.7\\
			\midrule
			\multicolumn{4}{l}{\textit{Integrated Combinations}}\\
			\quad Fuzzy Matching Combination & 74.9 & 80.1 & 69.7\\
			\quad All-Module Combination & 73.4 & 80.1 & 66.7\\
			\bottomrule
		\end{tabular*}
		\begin{tablenotes}
			\small
			\item Note: The ``Multidimensional Spatiotemporal Features'' include spatiotemporal energy distribution, spatiotemporal gradient fusion, and principal component analysis. The ``Extended Spatiotemporal Features'' further add spatiotemporal texture. The ``Best Dual Combination'' is spatiotemporal correlation + spatiotemporal texture; the ``Best Triple Combination'' adds depth-spectrum analysis; the ``Best Quadruple Combination'' further adds spatiotemporal energy distribution.
		\end{tablenotes}
		\label{tab:feature_modules}
	\end{threeparttable}
\end{table*}

This study adopted a detection rate threshold ($\geq$0.80) as the baseline constraint, with the loss multiplier for the generated class set to 1.008. As shown in Table I, results show that all individual modules and their combinations satisfied this requirement.Given the small variance in detection rate and its sensitivity to model bias, overall accuracy and real-sample detection rate were used as the primary performance indicators.
In the experimental design, single-module evaluation served as a form of ablation to examine the independent contribution of each feature. Results show that while single modules achieve only limited overall performance, they effectively reveal feature utility. Among them, spatiotemporal correlation and spatiotemporal texture achieved the best results, followed by depth-spectrum analysis, whereas energy distribution and principal component analysis were moderate, and other features contributed less.

Feature combination experiments demonstrated that multi-feature fusion substantially improves detection performance. The best-performing dual combination (spatiotemporal correlation + spatiotemporal texture) and triple combination (spatiotemporal correlation + spatiotemporal texture + depth-spectrum analysis) achieved overall accuracies of 76.3\% and 76.1\%, respectively, with real-sample detection rates exceeding 72\%. In importance-ranking strategies, the Top-80 and Top-100 feature subsets yielded the strongest results, reaching 74.6\% accuracy.

Strategic feature combinations generally underperformed compared to the optimized detector-based combinations. Only the statistical feature combination and PCA + correlation combination achieved relatively strong results, with accuracies of 73.4\% and 72.9\%, largely attributable to the presence of spatiotemporal correlation and texture features. Although multidimensional and extended spatiotemporal combinations theoretically integrate richer temporal-spatial cues, they did not outperform the more compact optimized combinations.

The fuzzy-matching combination achieved 74.9\% accuracy, ranking as the best-performing strategy outside of the optimized detector-based configurations.

Overall, these results validate the rationale of the experiment-driven feature subset exploration. Systematic evaluation confirmed spatiotemporal correlation, spatiotemporal texture, and depth-spectrum analysis as the three most effective feature types, and identified three high-performing combinations: the best-performing dual combination, the best-performing triple combination, and the fuzzy-matching combination. Notably, the all-feature combination, despite covering the full feature space, did not surpass these more compact strategies. This finding suggests that excessive feature inclusion introduces redundancy that may reduce discriminative efficiency, whereas careful feature selection and efficient combination can better balance sensitivity and real-sample preservation while maintaining the required detection rate threshold.

\subsubsection{Comparison with mainstream methods}
\begin{table*}[htbp]
	\centering
	\footnotesize                 
	\setlength{\tabcolsep}{2.8pt} 
	\begin{threeparttable}
		\caption{Performance of Different Models and Combinations across Video Generators (\%)}
		\begin{tabular*}{\textwidth}{@{\extracolsep{\fill}}l*{9}{c}@{}}
			\toprule
			Group/Model Combination & CogVideo & Mora & VideoCrafter2 & T2V-Zero & SVD & Modelscope & MuseV & HD-VG-130M(Real) & Overall Average \\
			\midrule
			Fuzzy Matching Combination (Ours) & 71.18 & 87.78 & 80.21 & 63.00 & 76.49 & 90.80 & 85.00 & 68.30 & 77.84 \\
			Best Dual Combination (Ours) & 74.77 & 82.94 & 87.51 & 63.20 & 71.31 & 89.71 & 82.30 & 72.92 & 78.08 \\
			Best Triple Combination (Ours) & 58.68 & 84.65 & 83.48 & 68.23 & 74.80 & 93.14 & 84.85 & 71.38 & 77.40 \\
			XCLIP & 66.81 & 62.55 & 93.73 & 77.19 & 16.77 & 34.33 & 24.59 & 90.34 & 58.29 \\
			XCLIP-Demamba & 61.58 & 69.82 & 94.37 & 65.18 & 16.74 & 50.29 & 28.23 & 86.34 & 59.07 \\
			ResNet50 & 29.04 & 64.47 & 79.25 & 45.71 & 18.52 & 7.44 & 88.41 & 99.04 & 54.01 \\
			AIGVDet & 25.57 & 53.86 & 71.30 & 19.40 & 11.34 & 1.28 & 86.83 & 99.04 & 46.08 \\
			VideoMAE & 22.41 & 27.93 & 61.86 & 0.16 & 27.30 & 26.41 & 35.42 & 98.82 & 37.54 \\
			MViT V2 & 38.50 & 58.98 & 67.96 & 59.64 & 13.34 & 21.02 & 31.72 & 96.20 & 48.42 \\
			\bottomrule
		\end{tabular*}
		\label{tab:model_comparisons_en}
	\end{threeparttable}
\end{table*}
Building upon the preceding experiments, we selected three high-performing configurations from DBINDS—namely, the Best Dual Combination (spatiotemporal correlation + spatiotemporal texture), the Best Triple Combination (including depth-spectrum analysis), and the Fuzzy-Matching Combination—subjected them to Bayesian hyperparameter optimization for further refinement, and compared their performance with existing detection methods and mainstream models. We do not report training‑set performance, since prior work \cite{GenVidbench} and our own runs indicate that high in‑generator accuracy is relatively easy to attain; the core question is generalization to unseen generators. Table II summarizes the cross‑method results under our one‑to‑many training protocol (training on Pika/Vript only; testing on seven unseen generators plus HD‑VG‑130M). 

On the unified test set, the three DBINDS configurations achieve comparable overall accuracy. The Best Dual Combination attains the highest accuracy on real videos while maintaining a high detection rate for generated videos; the Fuzzy‑Matching Combination ranks second; the Best Triple Combination is slightly lower. All three exhibit meaningful cross‑generator generalization. 

Subset‑level analysis reveals that the Best Triple Combination degrades on CogVideo, suggesting weaker separability of depth‑spectrum features for that generator, whereas its performance improves on Modelscope and T2V‑Zero. Conversely, the Fuzzy‑Matching Combination outperforms the other two on Mora, Stable Video Diffusion (SVD), and MuseV, indicating that a small fraction of heterogeneous features can be beneficial for certain generative mechanisms. Such variations are expected: architectural and data differences across generators produce model‑specific statistical “fingerprints,” so the discriminativeness of particular features can vary by source. 
\begin{figure}[!t]
	\centering
	\includegraphics[width=3.5in]{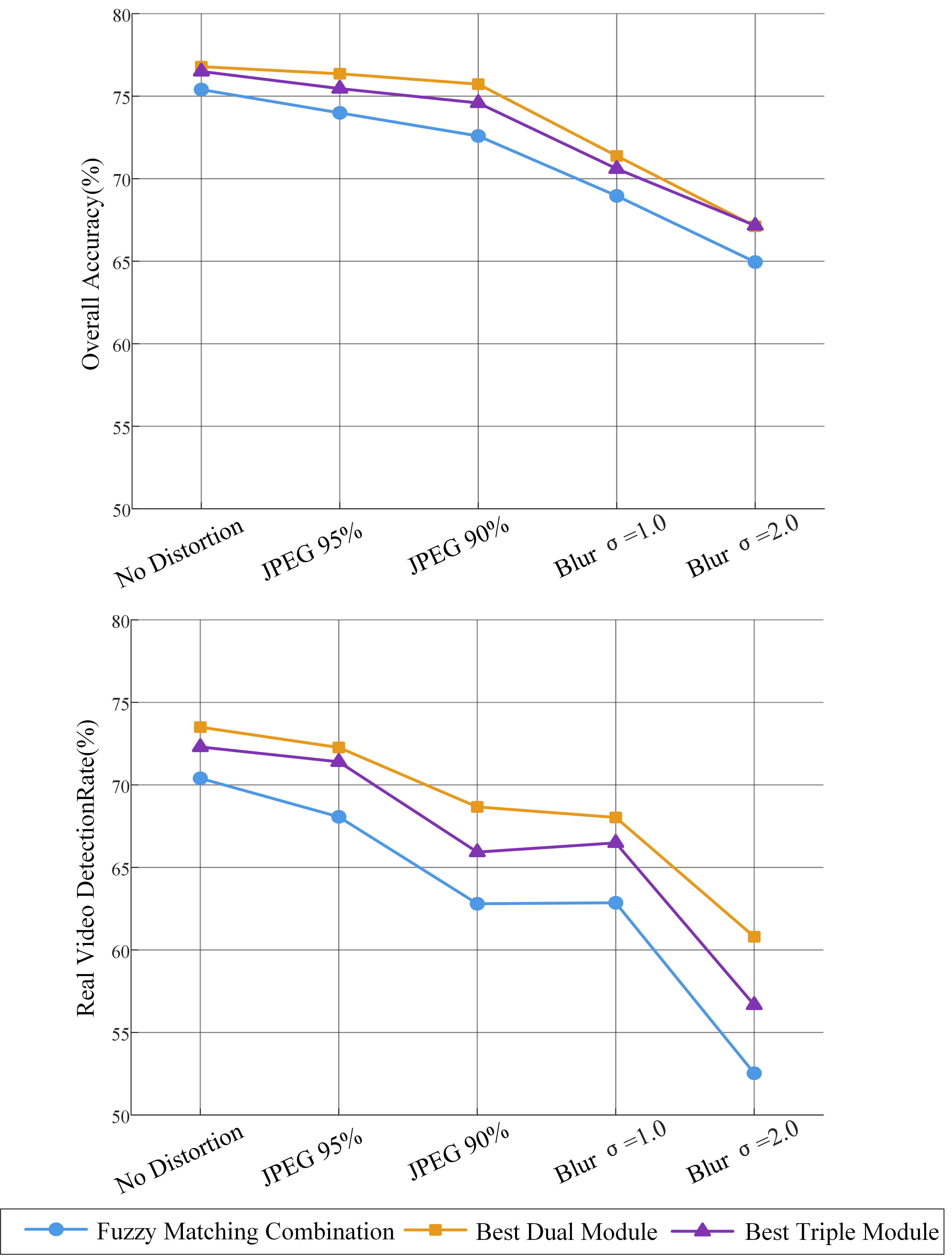}
	\caption{Trends in Model Performance under Various Perturbation Conditions}
	\label{fig_4}
\end{figure}

Compared with mainstream methods, the DBINDS configurations demonstrate stronger cross‑generator generalization in our small‑sample, one‑to‑many training regime. The classical CNN baseline (ResNet50) and large‑scale pretrained Transformers (XCLIP, VideoMAE, MViT V2) do not exhibit clear generalization advantages in this setting. DeMamba-XCLIP—known to generalize well at million‑scale training sets—performs among the best of the vision baselines here, though the margin to DBINDS is limited. Likely factors include the limited training data available in our setup and the substantial content diversity of GenVidBench (e.g., animation, game footage, static flora), which complicate learning stable visual cues. AIGVDet is illustrative: despite combining RGB and optical flow with a ResNet50 backbone, it underperforms the ResNet50 baseline under our protocol, suggesting that optical flow estimates can become noisy in heterogeneous scenarios. In contrast, DBINDS, trained on a single generator, captures regularities in latent‑space modeling and achieves higher detection rates and robustness on unseen samples. This advantage is particularly pronounced in few-shot scenarios. While methods such as XCLIP-Demamba demand millions of training samples and substantial memory resources, our approach requires only a single RTX 4090 GPU—or even lower hardware configurations—to complete training and still achieve a high detection rate, thereby significantly reducing the overall training cost.
\subsubsection{Robustness results}
\begin{table*}[htbp]
	\centering
	\footnotesize
	\caption{Model Performance Comparison under Different Perturbation Conditions (Transposed, Overall Accuracy, \%)}
	\begin{tabular}{lccc}
		\toprule
		Perturbation / Model & Fuzzy Matching Combination & Best Dual Combination & Best Triple Combination \\
		\midrule
		No Perturbation & Acc: 75.40 / Real: 70.40 & Acc: 76.79 / Real: 73.50 & Acc: 76.50 / Real: 72.30 \\
		JPEG 95\% & Acc: 73.99 / Real: 68.07 & Acc: 76.36 / Real: 72.27 & Acc: 75.46 / Real: 71.40 \\
		JPEG 90\% & Acc: 72.59 / Real: 62.80 & Acc: 75.73 / Real: 68.67 & Acc: 74.59 / Real: 65.93 \\
		Blur $\sigma=1.0$ & Acc: 68.96 / Real: 62.86 & Acc: 71.38 / Real: 68.03 & Acc: 70.60 / Real: 66.49 \\
		Blur $\sigma=2.0$ & Acc: 64.95 / Real: 52.53 & Acc: 67.12 / Real: 60.80 & Acc: 67.16 / Real: 56.67 \\
		Max Drop & 10.45 / 17.78 & 9.67 / 12.70 & 9.34 / 15.63 \\
		\bottomrule
	\end{tabular}
	\label{tab:perturbation_comparison_transposed}
\end{table*}

We assess the robustness of three strong DBINDS configurations—Fuzzy‑Matching Combination, Best Dual Combination, and Best Triple Combination (all with TPE‑optimized hyperparameters)—under common image perturbations. Specifically, we simulate JPEG compression (quality factors Q=95 and Q=90) and Gaussian blur ($\sigma$=1.0 and $\sigma$=2.0) to probe stability in practical scenarios. Table III and Fig. 4 report aggregate metrics and degradation curves. 

Under JPEG compression, all three methods remain relatively stable. The Best Dual Combination shows the smallest drop (0.43\% at Q=95), outperforming the Fuzzy‑Matching Combination (1.41\%); at Q=90, the drops are 1.06\% and 2.81\%, respectively, indicating stronger resistance to compression artifacts. 

Gaussian blur exerts a larger impact. With $\sigma$=1.0, overall accuracy decreases by roughly 5.4\%–6.4\% across methods; with $\sigma$=2.0, the decrease reaches 9.3\%–10.5\%. Even so, the Best Dual/Triple Combinations maintain higher absolute accuracy at the same blur level; for $\sigma$=2.0, they achieve 67.12\% and 67.16\%, respectively, compared with 64.95\% for the Fuzzy‑Matching Combination.

Metric‑level trends reveal an asymmetry: the Real‑Sample Preservation Rate degrades more than the Generated Detection Rate under perturbations. For example, at $\sigma$=2.0 the Fuzzy‑Matching Combination drops from 70.40\% to 52.53\% (-17.87 pp) in real‑sample preservation, whereas the Best Dual Combination drops from 73.50\% to 60.80\% (-12.70 pp). This suggests that perturbations more readily induce false positives on real videos—likely because natural textures are disrupted—while generator‑specific patterns remain partly detectable. 

These findings indicate that the quality and complementarity of feature combinations are critical for robustness. Although the Fuzzy‑Matching Combination aggregates a broader set of features, it is less stable under diverse perturbations than the more compact, complementary Best Dual/Triple Combinations. The Best Dual Combination, in particular, exhibits the most balanced robustness, especially against compression. Nonetheless, strong blur remains a principal challenge; in deployment, feature sets that preserve real‑sample reliability under varied transforms should be prioritized to balance detection performance and false‑positive risk. Overall, this study confirms that the Best Dual Combination (spatiotemporal correlation + spatiotemporal texture) is the most effective within DBINDS in terms of jointly maintaining accuracy and robustness. 
\begin{table}[htbp]
	\caption{Time Spent Processing 24,000 Frames (in Minutes)\label{tab:time_spent}}
	\centering
	\begin{tabular}{|c||c|}
		\hline
		\textbf{Number of Inversion Steps} & \textbf{Time (Minutes)} \\
		\hline
		1 & 47 \\
		\hline
		5 & 111 \\
		\hline
		10 & 183 \\
		\hline
		15 & 261 \\
		\hline
		20 & 359 \\
		\hline
	\end{tabular}
\end{table}
\subsubsection{Experimental Assessment of the Accuracy of Inversion}
\begin{table*}[htbp]
	\centering
	\footnotesize
	\setlength{\tabcolsep}{2.8pt}
	\begin{threeparttable}
		\caption{Performance of Different Feature Modules vs. Inversion Steps (Transposed, \%)}
		\begin{tabular*}{\textwidth}{@{\extracolsep{\fill}}l*{10}{c}@{}}
			\toprule
			Inversion Step &
			\makecell[c]{Spatiotemporal\\Energy} &
			\makecell[c]{Principal\\Component} &
			\makecell[c]{Spatiotemporal\\Correlation} &
			\makecell[c]{Spatiotemporal\\Gradient} &
			\makecell[c]{Wavelet\\Multiscale} &
			\makecell[c]{Depth\\Spectrum} &
			\makecell[c]{Spatiotemporal\\Local Var.} &
			\makecell[c]{Higher\\Statistics} &
			\makecell[c]{Spatiotemporal\\Texture} &
			\makecell[c]{Best Dual\\Comb.} \\
			\midrule
			20 & 61.2 & 58.8 & 65.9 & 60.4 & 56.7 & 63.1 & 59.0 & 57.3 & 62.8 & 67.1 \\
			15 & 64.2 & 55.8 & 66.6 & 62.3 & 60.7 & 65.2 & 61.7 & 60.4 & 66.0 & 71.5 \\
			10 & 64.2 & 63.5 & 72.7 & 61.0 & 55.3 & 65.7 & 61.5 & 58.0 & 71.2 & 76.3 \\
			5  & 63.1 & 61.2 & 75.1 & 60.6 & 56.1 & 64.7 & 59.8 & 57.2 & 68.4 & 77.2 \\
			1  & 61.0 & 54.4 & 54.5 & 55.4 & 55.7 & 62.0 & 57.7 & 52.9 & 58.1 & 59.2 \\
			\bottomrule
		\end{tabular*}
		\begin{tablenotes}
			\small
			\item Note: Accuracies are reported at five inversion-step settings (20, 15, 10, 5, 1) for each feature module/combination.
		\end{tablenotes}
		\label{tab:feature_inversion_transposed}
	\end{threeparttable}
\end{table*}
Table IV reports detection performance for individual feature modules and for the Best Dual Combination under different numbers of inversion steps. Contrary to the naive “more steps is better” assumption, the relationship between inversion steps and performance is clearly nonlinear. 
\begin{figure*}[!t]
	\centering
	\includegraphics[width=\textwidth]{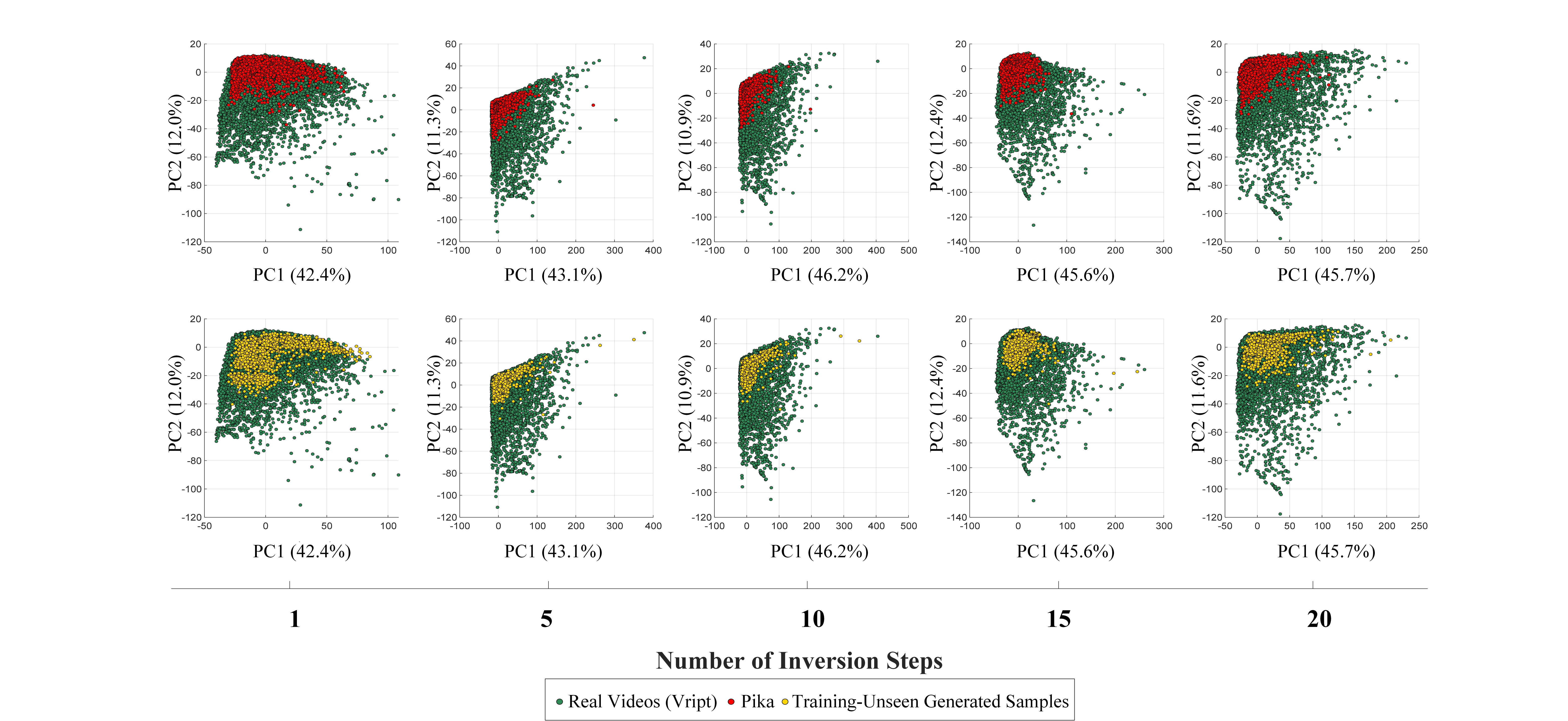}
	\caption{Visualization of the distribution changes in PCA based on the full feature combination at different inversion steps. The first row presents the scatter plots of the training samples Pika and Vript(real videos), while the second row shows the scatter plots of Vript and the seven unseen generated samples in the test set.}
	\label{final_fig}
\end{figure*}

With a small number of inversion steps (e.g., 1 step), all modules perform poorly; increasing the number of steps to 5 yields substantial gains. The spatiotemporal-correlation features reach 75.1\% overall accuracy with a 70.2\% real-sample detection rate, while the Best Dual Combination attains 77.2\% accuracy and 74.3\% preservation. Beyond that point (e.g., 10, 15, or 20 steps), most modules degrade. For the Best Dual Combination, overall accuracy drops from 77.2\% (with 5 steps) to 67.1\% (with 20 steps), and the real-sample detection rate declines from 74.3\% to 54.0\%. As illustrated in Fig. 5, excessive inversion tends to pull real samples toward the low-dimensional manifold learned by the generator, attenuating natural variability and details, which increases false positives on real videos; generated samples, already near that manifold, are comparatively less affected.

In terms of computational cost, increasing the number of inversion steps significantly increases the runtime. Table V shows the time required to process 3,000 videos (24,000 frames) on an RTX 4090. When the number of inversion steps is 1, it takes approximately 47 minutes, whereas increasing the steps to 20 results in a runtime of about 359 minutes (approximately 7.6 times longer). With 5 inversion steps, the processing time is around 111 minutes, while 10 steps require approximately 183 minutes, representing a nearly 40\% reduction in processing time. In balancing accuracy and computational efficiency, 5 inversion steps provide the best performance: it improves overall accuracy while significantly reducing preprocessing time.

In summary, the number of inversion steps should be selected based on a trade-off rather than being maximized. The evidence from the results indicates that 5 inversion steps provide the best cost-effectiveness in this study—even compared with the commonly used 10 steps—thereby offering practical guidance for DBINDS deployment and parameterization. These results indicate that higher inversion fidelity does not necessarily translate into better detection performance; instead, there exists a task‑dependent sweet spot in the step budget.

\section{Discussion}

\subsubsection{Inversion time overhead}Runtime is critical for deployment. Our experiments reveal a non‑monotonic relationship between the number of inversion steps and performance: too few steps do not establish stable discriminative cues, whereas too many lead to degradation and higher latency. Practical use therefore requires a balance between accuracy and time. Two directions follow: reduce the step budget required to achieve separability through adaptive step sizing, early stopping, or learned accelerators, thereby lowering computational cost; and introduce task‑aware constraints within the inversion loop to amplify separability in key latent cues—especially spatiotemporal correlation and spatiotemporal texture—so that accuracy improves even under limited step budgets. 

\subsubsection{Integration with existing methods} Empirical results and visualizations indicate persistent overlap between real and generated samples in latent space, reflecting the limits of current generators and real‑world content complexity. Future work may both discover and select features that specifically target the overlapping region and use DBINDS as an interpretable latent‑space branch within a multi‑stream system, integrated with end‑to‑end vision models and other modalities/metadata to improve overall performance and stability. 

\section{Conclusion}

This work advances AI‑generated video detection from the pixel to the latent domain via DBINDS. We derive an Initial Noise Difference Sequence (INDS) through diffusion inversion and analyze it with a multi‑dimensional, multi‑scale, multi‑domain framework, coupled with experiment‑driven feature optimization, LightGBM, and TPE. Under a strict one‑to‑many setting (single‑generator training; seven unseen generators plus one unseen real subset), DBINDS achieves 78.08\% overall accuracy, demonstrating strong generalization and robustness with modest training/deployment cost. Empirically, spatiotemporal correlation and spatiotemporal texture emerge as the most discriminative feature domains, guiding feature design and confirming INDS as an effective latent cue for practical deployment.We hope this latent‑space perspective provides a complementary pathway for future research and encourages continued exploration of latent representations for AI‑generated video detection.
\section*{Acknowledgments}
This work was supported in part by the National Natural Science Foundation of China under Grant 62562005, the Gansu Province University Faculty Innovation Fund Project under Grant 2025A-124, the Lanzhou City Science and Technology Program under Grant 2023-1-53, and the Anning District Science and Technology Program of Lanzhou under Grant 024-JB-5.
\bibliographystyle{IEEEtran}
\bibliography{reference}


\newpage

\vfill

\end{document}